\pdfoutput=1

\documentclass[11pt]{article}

\usepackage{acl}

\usepackage{times}
\usepackage{latexsym}

\usepackage[T1]{fontenc}

\usepackage[utf8]{inputenc}

\usepackage{microtype}

\usepackage{amssymb}
\usepackage{pifont}
\newcommand{\cmark}{\ding{51}}%
\newcommand{\xmark}{\ding{55}}%
\usepackage{graphicx}
\usepackage{multirow}
\usepackage{tabularx}

\title{TelME: Teacher-leading Multimodal Fusion Network for Emotion Recognition in Conversation}

\author{Taeyang Yun, Hyunkuk Lim, Jeonghwan Lee, Min Song\thanks{*Corresponding author} \\
Yonsei University, Seoul, South Korea \\
\textit{\{yuntaeyang0629, lsh950919, jeonghwan.ai, min.song\}@yonsei.ac.kr}
}

\begin{document}

  \maketitle
  
\begin{abstract}
Emotion Recognition in Conversation (ERC) plays a crucial role in enabling dialogue systems to effectively respond to user requests. The emotions in a conversation can be identified by the representations from various modalities, such as audio, visual, and text. However, due to the weak contribution of non-verbal modalities to recognize emotions, multimodal ERC has always been considered a challenging task. In this paper, we propose Teacher-leading Multimodal fusion network for ERC (TelME). TelME\footnote{Our code can be found here: \url{https://www.github.com/yuntaeyang/TelME}} incorporates cross-modal knowledge distillation to transfer information from a language model acting as the teacher to the non-verbal students, thereby optimizing the efficacy of the weak modalities. We then combine multimodal features using a shifting fusion approach in which student networks support the teacher. TelME achieves state-of-the-art performance in MELD, a multi-speaker conversation dataset for ERC. Finally, we demonstrate the effectiveness of our components through additional experiments. 

\end{abstract}

\section{Introduction}
\begin{figure}[t] 

\begin{center}

\includegraphics[width=1.0\linewidth]{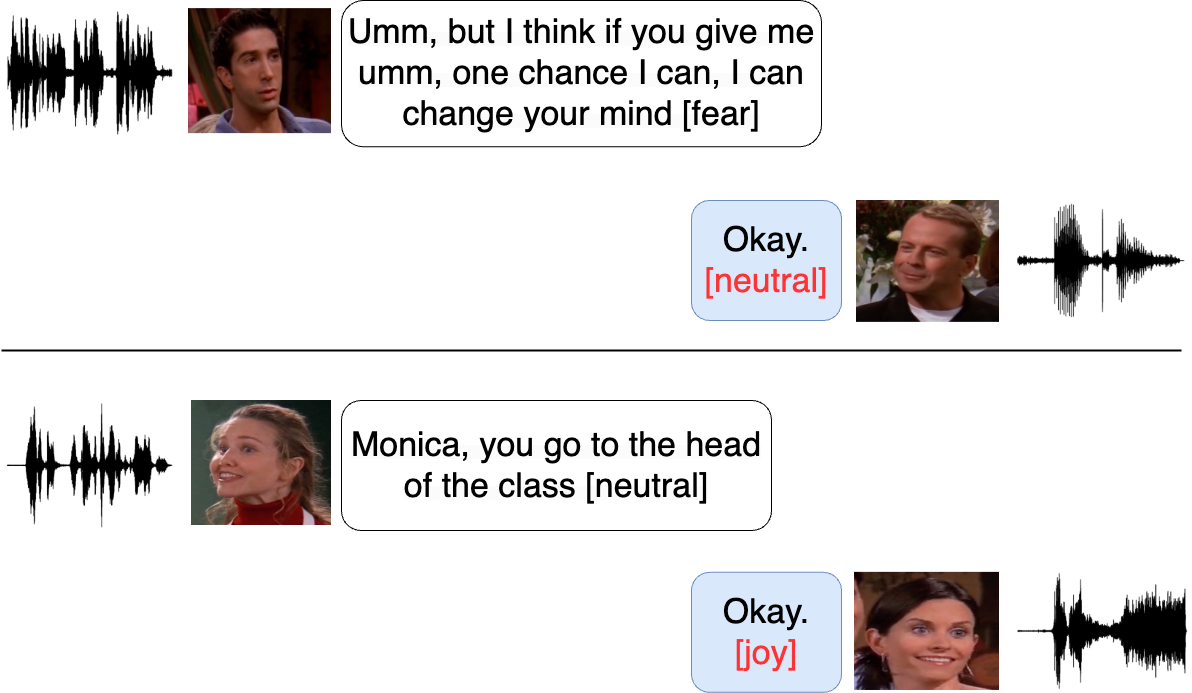}

\end{center}

\caption{Examples of multimodal ERC. Even the same "Okay" answer varies depending on the conversation situation and captures emotions in various modalities.}

\label{fig:figure_1}

\end{figure}

Emotion recognition holds paramount importance, enhancing the engagement of conversations by providing appropriate responses to the emotions of users in dialogue systems~\citep{ma2020survey}. The application of emotion recognition spans various domains, including chatbots, healthcare systems, and recommendation systems, demonstrating its versatility and potential to enhance a wide range of applications~\citep{poria2019emotion}. Emotion Recognition in Conversation (ERC) aims to identify emotions expressed by participants at each turn within a conversation. The dynamic emotions in a conversation can be detected through multiple modalities such as textual utterances, facial expressions, and acoustic signals~\citep{baltruvsaitis2018multimodal, liang2022foundations, majumder2019dialoguernn, hu2022unimse, chudasama2022m2fnet}. Figure \ref{fig:figure_1} illustrates an example of a multimodal ERC.

Much research on ERC has mainly focused on context modeling from text modality, disregarding the rich representations that can be obtained from audio and visual modalities. Text-based ERC methods have demonstrated that contextual information derived from text data is a powerful resource for emotion recognition~\citep{kim2021emoberta, lee2021compm, song2022supervised, song2022emotionflow}. However, non-verbal cues such as facial expressions and tone of voice, which are not covered by text-based methods, provide important information that needs to be explored in the field of ERC. Multimodal approaches demonstrate the possibility of integrating features from three modalities to improve the robustness of ERC systems \citep{mao2021dialoguetrm, chudasama2022m2fnet, hu2022unimse}. Nevertheless, these frameworks frequently ignore the varying degrees of impact the individual modalities have on emotion recognition and instead treat them as homogeneous components. This implies a promising opportunity to improve the ERC system by differentiating the level of contribution made by each modality.

\begin{figure}[] 

\begin{center}

\includegraphics[width=1.0\linewidth]{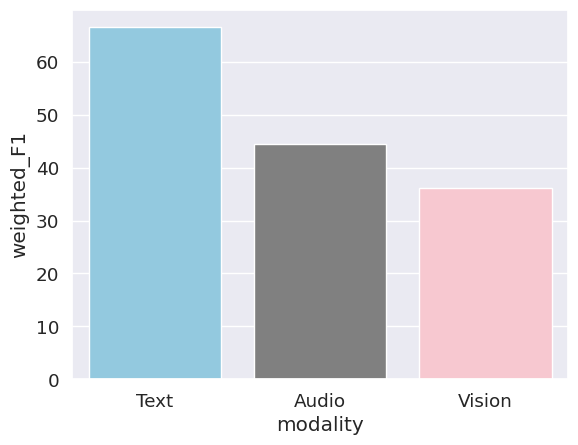}

\end{center}

\caption{Unimodal Performance on MELD dataset}

\label{fig:figure_2}

\end{figure}

In this paper, we propose Teacher-leading Multimodal fusion network for the ERC task (TelME) that strengthens and fuses multimodal information by accentuating the powerful modality while bolstering the weak modalities. Knowledge Distillation (KD) can be extended to transfer knowledge across modalities, where a powerful modality can play the role of a teacher to share knowledge with a weak modality~\citep{hinton2015distilling, xue2022modality}. While Figure \ref{fig:figure_2} shows the robustness of text in ERC tasks compared to the other two modalities, the other modalities present valuable information nonetheless. Thus, TelME enhances the representations of the two weak modalities through KD utilizing the text encoder as the teacher.
Our approach aims to mitigate heterogeneity between modalities while allowing students to learn the preferences of the teacher. TelME then incorporates Attention-based modality Shifting Fusion, where the student networks strengthened by the teacher at the distillation stage assist the robust teacher encoder in reverse, providing details that may not be present in the text. Specifically, our fusion method creates displacement vectors from non-verbal modalities, which are used to shift the emotion embeddings of the teacher.

We conduct experiments on two widely used benchmark datasets and compare our proposed method with existing ERC methods. Our results show that TelME performs well on both datasets and particularly excels in multi-party conversations, achieving state-of-the-art performance. 
The ablation study also demonstrates the effectiveness of our knowledge distillation strategy and its interaction with our fusion method.

Our contributions can be summarized as follows:

\begin{itemize}
    \item We propose Teacher-leading Multimodal fusion network for Emotion Recognition in Conversation (TelME). The proposed method considers different contributions of text and non-verbal modalities to emotion recognition for better prediction.

    \item To the best of our knowledge, we are the first to enhance the effectiveness of weak non-verbal modalities for the ERC task through cross-modal distillation.

    \item TelME shows comparable performance in two widely used benchmark datasets and especially achieves state-of-the-art in multi-party conversational scenarios.
\end{itemize}

\section{Related Work}
\subsection{Emotion Recognition in Conversation}
Recently, ERC has gained considerable attention in the field of emotion analysis. ERC can be categorized into text-based and multimodal methods, depending on the input format. Text-based methods primarily focus on context modeling and speaker relationships~\citep{jiao2019higru, li2020hitrans, hu2021dialoguecrn}. In recent studies~\citep{lee2021compm, song2022supervised}, context modeling has been carried out to enhance the understanding of contextual information by pre-trained language models using dialogue-level input compositions. Additionally, there are graph-based approaches~\citep{zhang2019modeling, ishiwatari2020relation, shen2021directed, ghosal2019dialoguegcn} and approaches that utilize external knowledge~\citep{zhong2019knowledge, ghosal2020cosmic, zhu2021topic}.
\begin{figure*}[hbt!] 

\begin{center}

\includegraphics[width=1.0\linewidth]{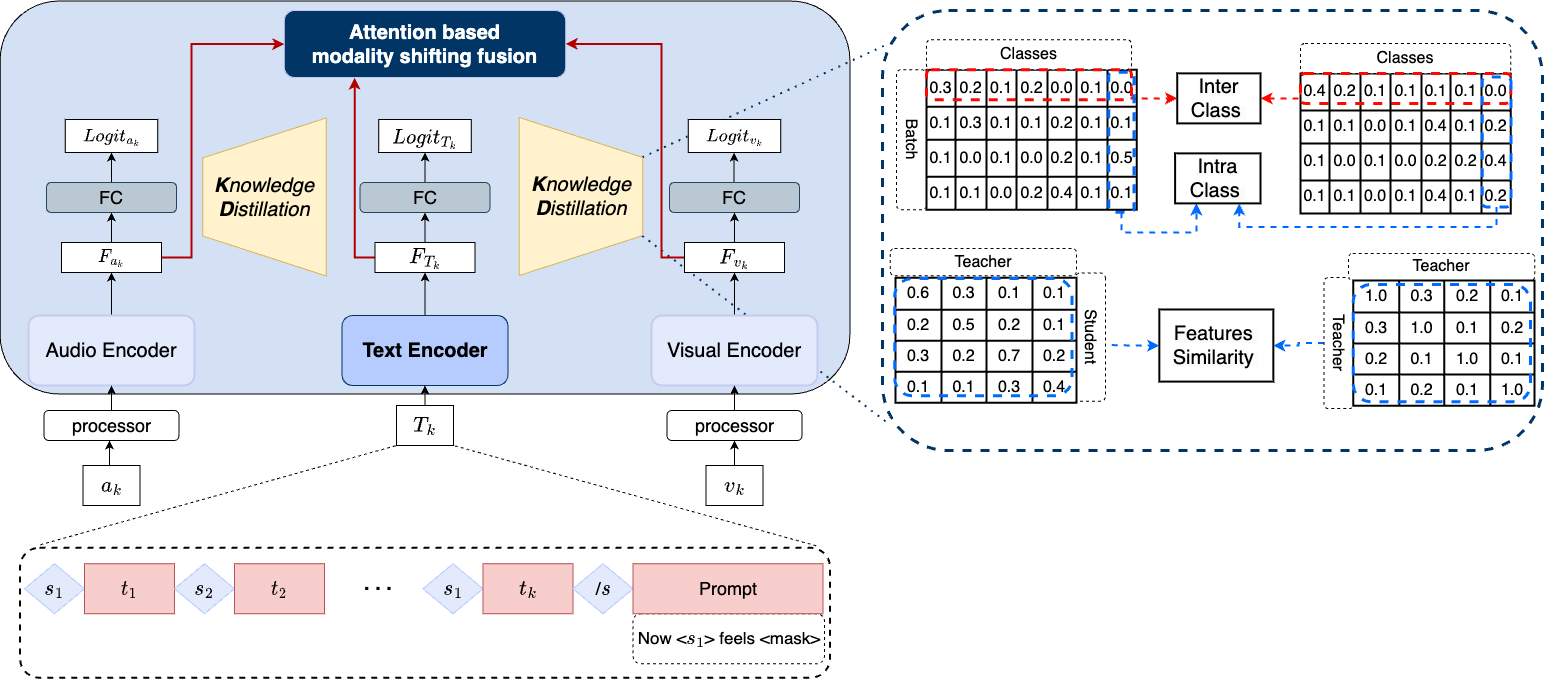}

\end{center}

\caption{The overview of TelME}

\label{fig:figure_3}

\end{figure*}

On the contrary, multimodal methods~\citep{poria2017context, hazarika2018icon, hazarika2018conversational, majumder2019dialoguernn} reflect dialogue-level multimodal features through recurrent neural network-based models. Other multimodal approaches~\citep{mao2021dialoguetrm, chudasama2022m2fnet} integrate and manipulate utterance-level features through hierarchical structures to extract dialogue-level features from each modality. EmoCaps~\citep{li2022emocaps} considers both multimodal information and contextual emotional tendencies to predict emotions. UniMSE~\citep{hu2022unimse} proposes a framework that leverages complementary information between Multimodal Sentiment Analysis and ERC. Unlike these methods, our proposed TelME is one in which the strong teacher leads emotion recognition while simultaneously bolstering attributes from weaker modalities to complement and enhance the teacher.

\subsection{Knowledge Distillation} 
The initial proposition of KD~\citep{hinton2015distilling} involves transferring knowledge by reducing the KL divergence between the prediction logits of teachers and students, demonstrating its effectiveness through improved performance of the student models. Subsequently, KD has been extended to distillation between intermediate features~\citep{heo2019comprehensive}. Furthermore, KD approaches~\citep{gupta2016cross, jin2021msd, tran2022within} have also been shown to transfer knowledge between modalities effectively in multimodal studies. ~\citet{li2023decoupled} mitigate multimodal heterogeneity by constructing dynamic graphs in which each vertex exhibits modality and each edge exhibits dynamic KD. However, since this work is not a study of ERC and is based on graph distillation, there is an intrinsic difference from our KD strategy.~\citet{ma2023transformer} proposes a transformer-based model utilizing self-distillation for ERC. Our proposed method, in contrast, uses response and feature-based distillation simultaneously to maximize the effectiveness of two other modalities by the teacher network based on text modality.

\section{Method}
\subsection{Problem Statement}
Given a set of conversation participants $S$, utterances $U$, and emotion labels $Y$, a conversation consisting of k utterances is represented as $[(s_i, u_1, y_1), (s_j,u_2,y_2,...,(s_i,u_k,y_k)]$, where $s_i, s_j\in S$ are the conversation participants. If $i$ = $j$, then $s_i$ and $s_j$ refer to the same speaker. $y_k\in Y$ is the emotion of the $k$-th utterance in a conversation, which belongs to one of the predefined emotion categories. Additionally, $u_k\in U$ is the $k$-th utterance. $u_k$ is provided in the format of a video clip, speech segment, and text transcript. i.e., $u_k=\{t_k, a_k, v_k\}$, where $\{t,a,v\}$ denotes a text transcript, speech segment, and a video clip. The objective of ERC is to predict $y_k$, the emotion corresponding to the $k$-th utterance in a conversation.

\subsection{TelME}
\subsubsection{Model Overview}
We propose Teacher-leading Multimodal fusion network for ERC (TelME), as illustrated in Figure \ref{fig:figure_3}. This framework is devised based on the hypothesis that by exploiting the varying levels of modality-specific contributions to emotion recognition, there is a potential to enhance the overall performance of an ERC system. Therefore, we introduce a strategic approach that focuses on accentuating the powerful modality while bolstering the weak modalities. We first extract powerful emotional representations through context modeling from text modality while capturing auditory and visual features of the current speaker from non-verbal modalities. However, due to the limited emotional recognition capability of audio and visual features as well as the heterogeneity between the modalities, effective multimodal interactions cannot be guaranteed \citep{zheng2022multi}. We thus mitigate the heterogeneity between modalities while maximizing the effectiveness of non-verbal modalities by distilling emotion-relevant knowledge of the teacher model into non-verbal students. We also use a fusion method in which strong emotional features from the teacher encoder are shifted by referring to representations of students strengthened in reverse. In the subsequent sections, we discuss the three components of TelME: Feature Extraction, Knowledge Distillation, and Attention-based modality Shifting Fusion.

\subsubsection{Feature Extraction}
Figure \ref{fig:figure_3} visually illustrates how each modality encoder receives its corresponding input to extract emotional features. In this section, we explain the methodologies employed to generate emotional features corresponding to each modality's input signals.

\textbf{Text}: Following previous research~\citep{lee2021compm, song2022supervised}, we conduct context modeling, considering all utterances from the inception of the conversation up to the k-th turn as the context. To handle speaker dependencies and differentiate between speakers, we represent speakers using the special token, $<s_i>$. Additionally, we construct the prompt, "Now $<s_i>$ feels  $<mask>$" to emphasize the emotion of the most recent speaker. We report the effect of the prompt in Appendix~\ref{sec:appendix_2}. The emotional features are derived from the embedding of the special token, $<mask>$. For our text encoder, we employ the modified Roberta~\citep{liu2019roberta}, which has exhibited its efficacy across various natural language processing tasks. We can extract emotional features from the text encoder as follows.
\begin{equation}
	C_k = [<s_i>,t_1,<s_j>,t_2,...,<s_i>,t_k]  
\end{equation}
\begin{equation}
	P_k = Now <s_i> feels <mask> 
\end{equation}
\begin{equation}
	F_{T_k} = TextEncoder(C_k </s> P_k)
\end{equation}
where $<s_i>$ is the special token indicating the speaker and $</s>$ is the separation token of Roberta. $F_{T_k} \in R^{1 \times d}$ is the embeddings of the mask token, $<mask>$ and d is the dimension of the encoder.

\textbf{Audio}: Self-supervised learning using Transformer has witnessed remarkable achievement, not only within the field of natural language processing but also in the realms of audio and video~\citep{bertasius2021space, baevski2022data2vec}. In line with this trend, we set the initial state of our audio encoder with data2vec~\citep{baevski2022data2vec}. To focus solely on the voice of the current speaker, we only utilize a speech segment of the k-th utterance, denoted as $a_k$. This speech segment is processed according to the pre-trained processor. The audio encoder then extracts emotional features from the processed input as follows.

\begin{equation}
	F_{a_k} = AudioEncoder(a_k)
\end{equation}
where $F_{a_k} \in R^{1 \times d}$  is the embeddings of $a_k$ and d is the dimension of the encoder.

\textbf{Visual}: Following the same reasoning as the audio modality, we configure the initial state of our visual encoder using Timesformer~\citep{bertasius2021space}. In order to concentrate exclusively on the facial expressions of the current speaker, we solely utilize a video clip of the k-th utterance, denoted as $v_k$. We extract the frames corresponding to the k-th utterance from the video and construct $v_k$ through image processing. The visual encoder then extracts emotional features from the processed input as follows.

\begin{equation}
	F_{v_k} = VisualEncoder(v_k)
\end{equation}
where $F_{v_k} \in R^{1 \times d}$ is the embedding of $v_k$ and d is the dimension of the encoder.

\subsubsection{Knowledge Distillation}
Addressing the challenge of heterogeneity between modalities and low emotional recognition contributions of non-verbal modalities holds great potential in facilitating satisfactory multimodal interactions~\citep{zheng2022multi}. Thus, we distill strong emotion-related knowledge of a language model that understands linguistic contexts, thereby augmenting the emotional features extracted from the other two modalities with comparatively lower contributions. We employ two distinct types of knowledge distillation concurrently: response and feature-based distillation. The overall loss for the student can be composed of the classification loss, response-based distillation loss, and feature-based distillation loss, i.e.,
\begin{equation} \label{eqn:student}
    L_{student} = L_{cls} + \alpha L_{response} + \beta L_{feature}
\end{equation}
where $\alpha$ and $\beta$ are the factors for balancing the losses.

\textbf{$L_{response}$} utilizes DIST~\citep{huang2022knowledge}, a technique originally used in image networks, as a cross-modal distillation for ERC. As shown in Figure~\ref{fig:figure_2}, effective knowledge distillation can be challenging due to the significant gap between the text modality and the other two modalities. Therefore, unlike conventional KD methods, we use a KD approach(\textbf{$L_{response}$}) that utilizes Pearson correlation coefficients instead of KL divergence as follows.

\begin{equation}
	d(\mu,\upsilon) = 1 -\rho(\mu,\upsilon)
\end{equation}
where $\rho(\mu,\upsilon)$ is the Pearson correlation coefficient between two probability vectors $\mu$ and $\upsilon$.

Specifically, \textbf{$L_{response}$} aims to distill preferences (relative rankings of predictions) by teachers through the correlations between teacher and student predictions, which can usefully perform knowledge distillation even in the extreme differences between teacher and student. We gather the predicted probability distributions for all instances within a batch and calculate the Pearson correlation coefficient between the teacher and student for inter-class and intra-class relations (Figure \ref{fig:figure_3}). Subsequently, we transfer the inter-class and intra-class relation to the student. The specific formulation of the response-based distillation can be described as follows.

\begin{equation}
	Y_{i,:}^{t} = softmax(Z_{i, : }^{t}/\tau)
\end{equation}

\begin{equation}
    Y_{i,:}^{s} = softmax(Z_{i, : }^{s}/\tau)
\end{equation}

\begin{equation}
	L_{inter} = \frac{\tau^2}{B}\sum_{i=1}^{B}d(Y_{i,:}^{s},Y_{i,:}^{t})
\end{equation}

\begin{equation}
	L_{intra} = \frac{\tau^2}{C}\sum_{j=1}^{C}d(Y_{:,j}^{s},Y_{:,j}^{t})
\end{equation}

\begin{equation} \label{eqn: response}
	L_{response} = L_{inter} + L_{intra}
\end{equation}
Given a training batch $B$ and the emotion categories $C$, $Z^s \in R^{B \times C}$ is the prediction matrix of the student and $Z^t \in R^{B \times C}$ is the prediction matrix of the teacher. $\tau$ > 0 is a temperature parameter to control the softness of logits.

However, rather than relying solely on $L_{response}$, we introduce $L_{feature}$ as an additional distillation loss to better leverage the embedded information in the teacher network. $L_{feature}$ aims to mitigate the heterogeneity between the representations of the teacher and student models, allowing us to distill richer knowledge from the teacher compared to using only $L_{response}$. Through this, the features of the students can faithfully support the teacher during the multimodal fusion stage. \textbf{$L_{feature}$} leverages the similarity among normalized representation vectors of the teacher and the student within a batch (Figure~\ref{fig:figure_3}). We construct the target similarity matrix by performing a dot product between the representation matrix of the teacher and its transposition matrix. By applying the softmax function to this matrix, we derive the target probability distribution as follows. 
\begin{figure}[t] 

\begin{center}

\includegraphics[width=0.7\linewidth]{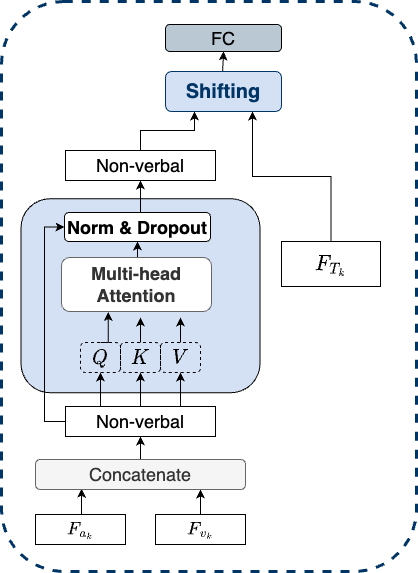}

\end{center}

\caption{Attention-based modality Shifting Fusion}

\label{fig:figure_4}

\end{figure}

\begin{equation}
	P_i = \frac{exp(M_{i,j} /\tau)}{\sum_{l= 1}^{B}exp(M_{i,l} /\tau)},  \forall  i,j \in B 
\end{equation}
where $B$ is a training batch and $M \in R^{B \times B}$ is the target similarity matrix. $\tau$ > 0 is a temperature parameter controlling the smoothness of the distribution. $P_i$ is the target probability distribution.

Similarly, we can compute the similarity matrix between the teacher and the student by taking the dot product of their representations. Subsequently, we can calculate the similarity probability distribution as follows. 

\begin{equation}
	Q_i = \frac{exp(M'_{i,j} /\tau)}{\sum_{l= 1}^{B}exp(M'_{i,l} /\tau)},  \forall  i,j \in B 
\end{equation}
where $M' \in R^{B \times B}$ is the similarity matrix of student and teacher. $Q_i$ is the similarity probability distribution of teacher and student.

With these two probability distributions, we compute the KL divergence as the loss for the feature-based distillation.

\begin{equation} \label{eqn: feature}
	L_{feature} = \frac{1}{B}\sum_{i=1}^{B} KL(P_i\parallel Q_i)
\end{equation}
where $KL$ is the Kullback–Leibler divergence.

\subsubsection{Attention-based modality Shifting Fusion}
The emotional features from the enhanced student networks can impact the teacher model's emotion-relevant representations, providing information that may not be captured from the text. To fully utilize these features, we adopt a multimodal fusion approach where feature vectors from the student models manipulate the representation vectors from the teacher, effectively incorporating non-verbal information into the representation vector. To highlight non-verbal characteristics, we concatenate the vectors of the student models and perform multi-head self-attention. The vectors of non-verbal information generated through the multi-head self-attention process and emotional features of the teacher encoder enter the input of the shifting step (Figure \ref{fig:figure_4}).  We are inspired by ~\citet{rahman2020integrating} to construct the shifting step. In the shifting step, a gating vector is generated by concatenating and transforming the vector of the teacher model and the vector of the non-verbal information. 
\begin{equation}
    g_{AV}^k = R(W_1 \cdot <F_{T_k}, F_{attention}^k> + b_1)
\end{equation}
where <,> is the operation of vector concatenation, $R(x)$ is a non-linear activation function, $W_1$ is the weight matrix for linear transform, and $b_1$ is scalar bias. $F_{attention}$ is the emotional representation vectors of non-verbal information. $g_{AV}$ is the gating vector. The gating vector highlights the relevant information in the non-verbal vector according to the representations of the teacher model. 
We define the displacement vector by applying the gating vector as follows.
\begin{equation}
H_k = g_{AV}^k \cdot (W_2 \cdot F_{attention}^k + b_2)
\end{equation}
where $W_2$ is the weight matrix for linear transform and $b_2$ is scalar bias. $H $ is the non-verbal information-based displacement vector. 

We subsequently utilize the weighted sum between the representation vector of the teacher and the displacement vector to generate a multimodal vector. Finally, we predict emotions using the multimodal vector.
\begin{equation}
Z_k = F_{T_k} + \lambda \cdot H_k
\end{equation}
\begin{equation}
\lambda = min(\frac{\|F_k \|_2}{\| H_k \|_2} \cdot \theta, 1)
\end{equation}
where $Z$ is the multimodal vector. We apply the scaling factor $\lambda$ to control the magnitude of the displacement vector and $\theta$ as a threshold hyperparameter. $\| F_k \|_2, \| H_k \|_2 $ denote the L2 norm of the  $F_k$ and $H_k$ vectors respectively.

\begin{table}[]
\resizebox{\columnwidth}{!}{%
\begin{tabular}{ccccccc}
\hline
\multirow{2}{*}{Dataset} & \multicolumn{3}{c}{IEMOCAP} & \multicolumn{3}{c}{MELD} \\
                         & train    & dev    & test    & train   & dev    & test  \\ \hline
Dialogue                 & 108      & 12     & 31      & 1038    & 114    & 280   \\
Utterance                & 5163     & 647    & 1623    & 9989    & 1109   & 2610  \\
Classes                    & \multicolumn{3}{c}{6}       & \multicolumn{3}{c}{7}    \\ \hline
\end{tabular}%
}
\caption{Statistics of the two benchmark datasets.}
\label{tab:Table1}
\end{table}

\section{Experiments}
\subsection{Datasets}
We evaluate our proposed network on MELD~\citep{poria2018meld} and IEMOCAP~\citep{busso2008iemocap} following other works on ERC listed in Appendix~\ref{sec:appendix_4}. The statistics are shown in Table ~\ref{tab:Table1}.

\textbf{MELD} is a multi-party dataset comprising over 1400 dialogues and over 13,000 utterances extracted from the TV series Friends. This dataset contains seven emotion categories for each utterance: neutral, surprise, fear, sadness, joy, disgust, and anger. 

\textbf{IEMOCAP} consists of 7433 utterances and 151 dialogues in 5 sessions, each involving two speakers per session. Each utterance is labeled as one of six emotional categories: happy, sad, angry, excited, frustrated and neutral. The train and development datasets consist of the first four sessions randomly divided at a 9:1 ratio. The test dataset consists of the last session.

We purposely exclude CMU-MOSEI~\citep{bagher-zadeh-etal-2018-multimodal}, a well-known multimodal sentiment analysis dataset, as it comprises single-speaker videos and is not suitable for ERC, where emotions dynamically change within each conversation turn.

\begin{table*}[]
\centering
\resizebox{\textwidth}{!}{%
\begin{tabular}{cccccccccc}
\hline
\multirow{2}{*}{Models} & \multicolumn{8}{c}{MELD: Emotion Categories}                           & IEMOCAP \\
                        & Neutral & Surprise & Fear  & Sadness & Joy   & Disgust & Anger & F1    & F1      \\ \hline
DialogueRNN~\citep{majumder2019dialoguernn}             & 73.50   & 49.40    & 1.20  & 23.80   & 50.70 & 1.70    & 41.50 & 57.03 & 62.75   \\
ConGCN~\citep{zhang2019modeling}                  & 76.70   & 50.30    & 8.70  & 28.50   & 53.10 & 10.60   & 46.80 & 59.40 & 64.18   \\
MMGCN~\citep{hu2021mmgcn}                   & -       & -        & -     & -       & -     & -       & -     & 58.65 & 66.22   \\
DialogueTRM~\citep{mao2021dialoguetrm}             & -       & -        & -     & -       & -     & -       & -     & 63.50 & 69.23   \\
DAG-ERC~\citep{shen2021directed}                 & -       & -        & -     & -       & -     & -       & -     & 63.65 & 68.03   \\
MM-DFN~\citep{hu2022mm}                  & 77.76   & 50.69    & -     & 22.94   & 54.78 & -       & 47.82 & 59.46 & 68.18   \\
M2FNet~\citep{chudasama2022m2fnet}                  & -       & -        & -     & -       & -     & -       & -     & 66.71 & 69.86   \\
EmoCaps~\citep{li2022emocaps}                 & 77.12   & \textbf{63.19}    & 3.03  & 42.52   & 57.50 & 7.69    & \textbf{57.54} & 64.00 & \textbf{71.77}   \\
UniMSE~\citep{hu2022unimse}                  & -       & -        & -     & -       & -     & -       & -     & 65.51 & 70.66   \\
GA2MIF~\citep{li2023ga2mif}                  & 76.92   & 49.08    & -     & 27.18   & 51.87 & -       & 48.52 & 58.94 & 70.00   \\
FacialMMT~\citep{zheng2023facial}               & 80.13   & 59.63    & 19.18 & 41.99   & 64.88 & 18.18   & 56.00 & 66.58 & -       \\ \hline \hline
\textbf{TelME}                   & \textbf{80.22}   & 60.33    & \textbf{26.97} & \textbf{43.45}   & \textbf{65.67} & \textbf{26.42}   & 56.70 & \textbf{67.37} & 70.48   \\ \hline
\end{tabular}
}
\caption{Performance comparisons on MELD (7-way) and IEMOCAP}
\label{tab:Table2}
\end{table*}

\subsection{Experiment Settings}
We evaluate all experiments using the weighted average F1 score on two class-imbalanced datasets. We use the initial weight of the pre-trained models from Huggingface’s Transformers~\citep{wolf2019huggingface}. The output dimension of all encoders is unified to 768. The optimizer is AdamW and the initial learning rate is 1e-5. We use a \emph{linear schedule with warmup} for the learning rate scheduler. All experiments are conducted on a single NVIDIA GeForce RTX 3090. More details are in Appendix~\ref{sec:appendix_3}.

\subsection{Main Results}
We compare TelME with various multimodal-based ERC methods (explained in Appendix~\ref{sec:appendix_4}) on both datasets in Table~\ref{tab:Table2}. TelME demonstrates robust results in both datasets and achieves state-of-the-art performance on MELD. Specifically, TelME outperforms the previous state-of-the-art method (M2FNet) in MELD by $0.66\%$, and exhibits a substantial $3.37\%$ improvement in MELD compared to EmoCaps, which currently achieves state-of-the-art performance in IEMOCAP. Previous methods, such as EmoCaps and UniMSE, have also shown effectiveness in IEMOCAP but exhibit somewhat weaker performance on MELD.

As shown in Table~\ref{tab:Table2}, we report the performance of various methods for emotion labels in MELD. TelME outperforms other models in all emotions except Surprise and Anger. However, assuming that Surprise and Fear, as well as Disgust and Anger, are similar emotions, Emocaps shows a bias towards Surprise and Anger during inference, only achieving $3.03\%$ and $7.69\%$ in F1 score for Fear and Disgust, respectively. On the other hand, TelME distinguishes these similar emotions better, bringing the scores for Fear and Disgust up to $26.97\%$ and $26.42\%$. We speculate that our framework predicts minority emotions more accurately as the non-verbal modality information (e.g., intensity and pitch of an utterance) enhanced through our KD strategy better assists the teacher in judging the confusing emotions.

\subsection{The Impact of Each Modality}

Table \ref{tab:Table4} presents the results for single-modality and multimodal combinations. The single-modality performances for audio and visual are the results after applying our knowledge distillation method, and the same fusion approach as TelME is used for dual-modality results. The text modality performs the best among the single-modality, which supports our decision to use the text encoder as the teacher model. Additionally, the combination of non-verbal modalities and text modality achieves superior performance compared to using only text. Our findings indicate that the audio modality significantly contributes more to emotion recognition and holds greater importance compared to the visual modality. We speculate this can be attributed to its ability to capture the intensity of emotion through variations in the tone and pitch of the speaker. Overall, our method achieves $3.52\%$ improvement in IEMOCAP and $0.8\%$ in MELD over using only text.

\begin{table}[]
\resizebox{\columnwidth}{!}{%
\begin{tabular}{cccc}
\hline
Methods                & Remarks      & IEMOCAP        & MELD  \\ \hline
Audio                 & KD           & 48.11         & 46.60 \\
Visual                & KD           & 18.85          & 36.72          \\
Text                  & -            & 66.60          & 66.57          \\
Text + Visual         & ASF          & 67.94          & 67.05          \\
Text + Audio          & ASF          & 69.26          & 67.19          \\ \hline
\textbf{TelME} &          & \textbf{70.48} & \textbf{67.37}          \\ \hline
\end{tabular}%
}
\caption{Performance comparison for single modality and multiple multimodal combinations}
\label{tab:Table4}
\end{table}

\subsection{Ablation Study}

\begin{table}[]
\resizebox{\columnwidth}{!}{%
\begin{tabular}{ccccc}
\hline
Dataset                  & ASF & L\_response & L\_feature & F1             \\ \hline
\multirow{4}{*}{IEMOCAP} & \xmark   & \xmark           & \xmark          & 63.33          \\
                         & \cmark   & \xmark           & \xmark          & 68.19          \\
                         & \cmark   & \cmark           & \xmark          & 69.42          \\
                         & \cmark   & \cmark           & \cmark          & \textbf{70.48}          \\ \hline
\multirow{4}{*}{MELD}    & \xmark   & \xmark           & \xmark          & 67.04          \\
                         & \cmark   & \xmark           & \xmark          & 66.75          \\
                         & \cmark   & \cmark           & \xmark          & 67.23          \\
                         & \cmark   & \cmark           & \cmark          & \textbf{67.37} \\ \hline
\end{tabular}%
}
\caption{Results of ablation study. Here, $L_{response}$ is our response-based distillation,  $L_{feature}$ is our feature-based distillation and ASF is our fusion method.}
\label{tab:Table3}
\end{table}

We conduct an ablation study to validate our knowledge distillation and fusion strategies in Table \ref{tab:Table3}.
The initial row for each dataset represents the outcome of training each modality encoder using cross-entropy loss and concatenating the embeddings without incorporating distillation loss and our fusion method.

Using our fusion method alone, IEMOCAP showed performance improvement, but MELD showed poor performance. The effectiveness of our fusion method in achieving optimal modality interaction cannot be guaranteed without knowledge distillation. Because each encoder is trained independently, focusing solely on improving its performance without considering the multimodal interaction. On the other hand, as our knowledge distillation components are added, these bring about consistent improvements for both datasets.

When we examine the specific effects of the KD strategy, we observe performance improvements for both datasets, even when using only $L_{response}$, presenting its efficacy in closing the gap between the teacher and the students. Furthermore, adding $L_{feature}$ aimed to leverage the richer knowledge of the teacher is more effective in IEMOCAP and shows marginal performance enhancements in MELD. However, we speculate that the slight improvement in MELD may be attributed to class imbalance.  While TelME significantly outperforms existing approaches in minority classes of MELD, the weighted F1 score is only slightly improved due to the low number of samples. We show an analysis of this class imbalance problem in Section~\ref{sec:appendix_5} as well as an error analysis of the emotion classes in Appendix~\ref{sec:appendix_6}.

Figure \ref{fig:figure_5} shows the individual performance of the audio and visual modalities based on the distillation loss. We observe that applying both types of distillation loss is more effective compared to not applying them. The performance of visual modality on the IEMOCAP dataset has declined, possibly because facial expressions are not effectively captured in the limited image frames of a short utterance. However, even with lower individual performance, all modalities have been shown to contribute to the improvement of emotion recognition performance through our approach (Table~\ref{tab:Table4}).
 

\begin{figure}[t!] 
\centering

\begin{center}

\includegraphics[width=1.0\linewidth]{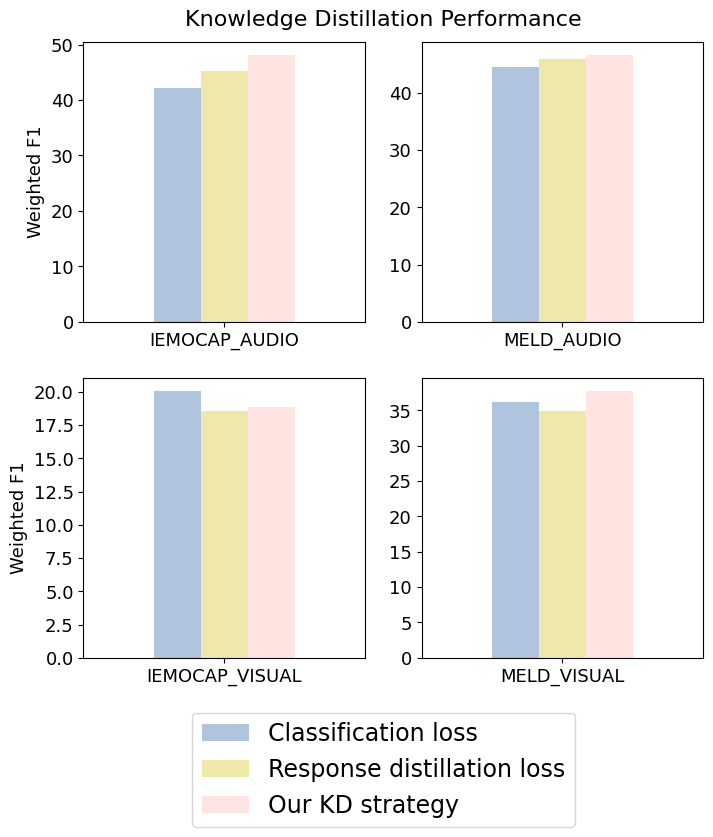}

\end{center}

\caption{Individual performance of audio and visual modalities according to knowledge distillation type.}

\label{fig:figure_5}
\end{figure}

\subsection{Study on Teacher Modality}
\label{sec:appendix_1}

\begin{table}[hbt!]
\centering
\resizebox{\columnwidth}{!}{%
\begin{tabular}{ccc}
\hline
                       & MELD  & IEMOCAP \\ \hline
TelME (Audio Teacher)  & 56.28 & 49.36   \\
TelME (Visual Teacher) & 56.85 & 56.78   \\ \hline
TelME (Text Teacher)   & \textbf{67.37} & \textbf{70.48}   \\ \hline
\end{tabular}%
}
\caption{TelME Performance by Teacher Modality}
\label{tab:appendix_A1}
\end{table}

To assess the optimality of employing text modality as the teacher, we conduct comparative experiments by setting each modality as the teacher modality, which is shown in Table~\ref{tab:appendix_A1}. Our study shows that the TelME framework performs best with the text encoder as the teacher, while treating the other modalities as the teacher significantly hinders model performance.

\begin{table}[hbt!]
\resizebox{\columnwidth}{!}{%
\begin{tabular}{ccccc}
\hline
 & & Audio Student & Visual Student & Text Student \\ \hline
& Audio Teacher  & 44.55 & 34.86 & 54.83 \\
MELD & Visual Teacher & 40.18 & 36.14 & 59.72 \\ 
& Text Teacher   & \textbf{46.60} & \textbf{36.72} & \textbf{66.60} \\\hline
& Audio Teacher   & 42.24    &  20.45    &  57.42  \\
IEMOCAP & Visual Teacher  & 44.13    & \textbf{22.06} & 63.94  \\ 
& Text Teacher    & \textbf{48.11}    &  18.85 & \textbf{66.57}
\\\hline
\end{tabular}%
}
\caption{Teacher Modality Study on MELD and IEMOCAP}
\label{tab:appendix_A1_meld}
\end{table}


Additionally, Table~\ref{tab:appendix_A1_meld} reports the individual performance of the student models based on the teacher modality. The diagonals (cases where the teacher and student modalities are the same) in Table~\ref{tab:appendix_A1_meld} represent results without performing Knowledge Distillation (KD). Our comparative experiment results show that a robust text encoder can most effectively serve as the teacher. Specifically, designating the text encoder as the teacher enhances the performance of all student models except for the visual student in IEMOCAP. On the other hand, it is evident that treating a weak non-verbal model as the teacher impairs student performance, thereby performing suboptimally compared to having a text-based teacher. 

\subsection{Class Imbalance}
\label{sec:appendix_5}
\begin{figure}[hbt!] 
\centering

\begin{center}

\includegraphics[width=1.0\linewidth]{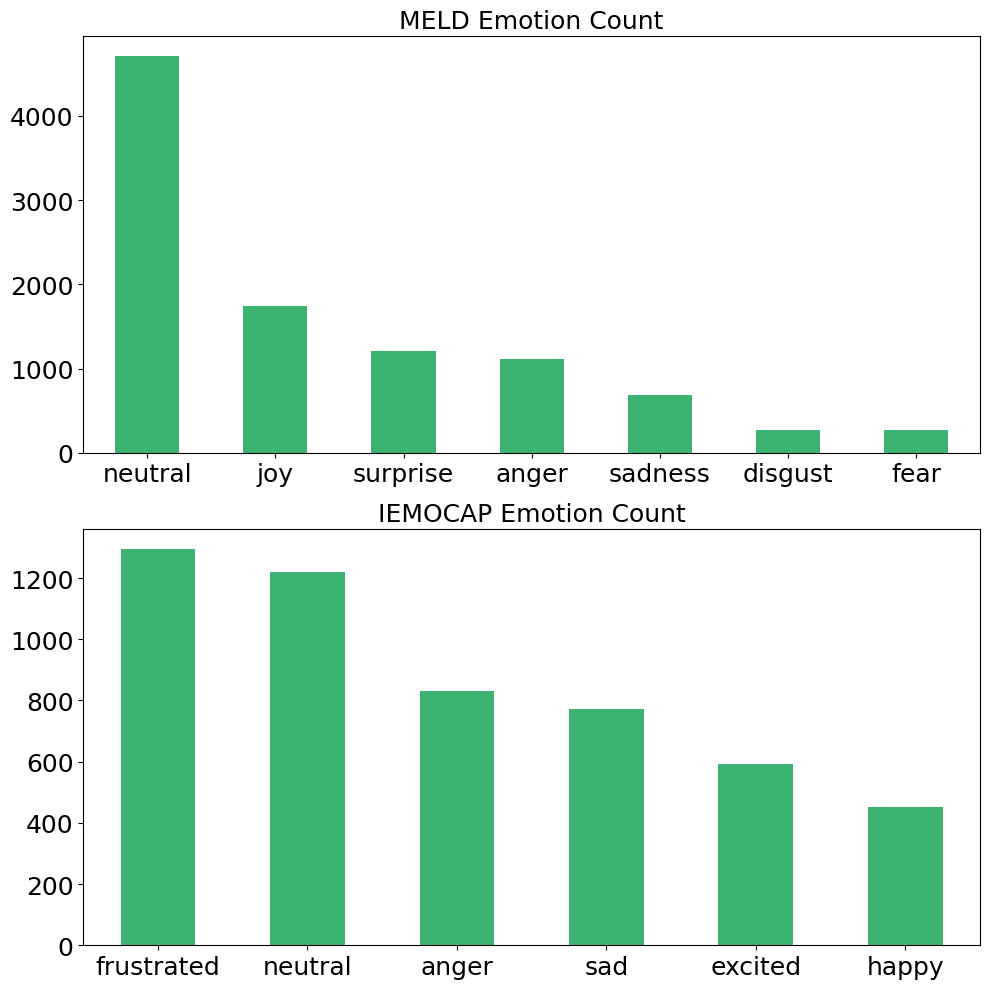}

\end{center}

\caption{Count distribution of emotion classes for both MELD and IEMOCAP datasets}

\label{fig:appendix_A5}
\end{figure}

Figure~\ref{fig:appendix_A5} illustrates the label distribution within the MELD and IEMOCAP datasets. Notably, the MELD dataset exhibits a pronounced imbalance, with the "neutral" class comprising the majority at $47\%$ of the data, followed by "joy" with $17\%$ and "surprise" with $12\%$. This substantial class imbalance presents a challenge in the context of distillation, specifically for the teacher encoder to initially identify the minority classes and subsequently transfer this information to the non-verbal student encoders. We believe that this class imbalance is a contributing factor to the limited observed improvements associated with $L_{feature}$ in the MELD dataset compared to the IEMOCAP dataset.

\section{Conclusion}
This paper proposes Teacher-leading Multimodal fusion network for ERC (TelME), a novel multimodal ERC framework. TelME incorporates a cross-modal distillation that transfers the knowledge of text encoders trained in linguistic contexts to enhance the effectiveness of non-verbal modalities. Moreover, we employ the fusion method that shifts the features of the teacher model by referring to non-verbal information. We show through experiments on two benchmarks that our approach is practical in ERC. TelME delivers robust performance in both datasets and especially achieves state-of-the-art results in the MELD dataset, which consists of multi-party conversational scenarios. We believe that this research presents a new direction that can incorporate multimodal information for ERC.

\section*{Limitations}
This study has a limitation wherein the visual modality shows a lower capability to recognize emotions compared to the audio modality. To address this limitation, future research should focus on developing techniques to accurately capture and interpret the facial expressions of the speaker during brief utterances. By improving the extraction of visual features, the effectiveness of knowledge distillation can be significantly enhanced, thus showcasing its potential to make a more substantial contribution to emotion recognition.

\section*{Acknowledgements}
This work was partly supported by Institute of Information \& Communications Technology Planning \& Evaluation (IITP) Grant funded by the Korea government (MSIT), Artificial Intelligence Graduate School Program, Yonsei University, under Grant 2020-0-01361 and the National Research Foundation of Korea(NRF) grant funded by the Korea government(MSIT) (No. 2022R1A2B5B02002359).

\bibliography{anthology, custom}
\bibliographystyle{acl_natbib}

\appendix

\section{Appendix}
\label{sec:appendix}

\subsection{Effect of the prompt}
\label{sec:appendix_2}
\begin{table}[hbt!]
\centering
\resizebox{\columnwidth}{!}{%
\begin{tabular}{ccc}
\hline
                 & MELD  & IEMOCAP \\ \hline
w/o prompt ([cls]+context)       & 65.25 & 66.48   \\
context + prompt & \textbf{66.57} & \textbf{66.60}   \\ \hline
\end{tabular}%
}
\caption{Comparison of the teacher performance based on the use of the prompt}
\label{tab:appendix_A2}
\end{table}
Table~\ref{tab:appendix_A2} shows an ablation experiment on the prompt. We remove the prompt and use the CLS token to compare emotion prediction results with the results using the prompt. We observe from the results that the prompt helps to infer the emotion of a recent speaker from a set of textual utterances. 

\subsection{Hyperparameter Settings}
\label{sec:appendix_3}
\begin{table}[hbt!]
\resizebox{\columnwidth}{!}{%
\begin{tabular}{lcc}
\hline
\multicolumn{1}{c}{Hyperparameter}           & IEMOCAP & MELD \\ \hline
\multicolumn{3}{l}{Knowledge distillation}                    \\ \hline
Balance factors for $L_{student}$               & $\alpha$=0.1       & 1    \\
Temperature for $L_{response}$                  & 4       & 2    \\
Temperature for $L_{feature}$                   & 1       & 1    \\ \hline
\multicolumn{3}{l}{Attention modality Shifting Fusion}        \\ \hline
Threshold parameter                          & 0.01     & 0.1  \\
Dropout                                      & 0.2     & 0.1  \\
The number of heads for multi-head attention & 4       & 3    \\ \hline
\end{tabular}%
}
\caption{hyperparameter settings of TelME on two datasets}
\label{tab:appendix_A3}
\end{table}

Through our KD strategy, audio and visual encoders are trained using the loss functions mentioned in Equation \ref{eqn:student}. In $L_{student}$, the balancing factors are all set to 1, excluding $\alpha$ for IEMOCAP. The temperature parameter for the $L_{response}$ function is adjusted to 4 for MELD and 2 for IEMOCAP. The temperature parameter for $L_{feature}$ is set to 1 regardless of the dataset. We also use a fusion method that shifts vectors in the teacher model, where the threshold parameter is set to 0.01 for IEMOCAP and 0.1 for MELD. Furthermore, Dropout is adjusted to 0.2 for MELD and 0.1 for IEMOCAP. The number of heads used in the multi-head attention process is 4 for IEMOCAP and 3 for MELD.

\subsection{Compared Models}
\label{sec:appendix_4}
We compare TelME against the following models: DialogueRNN \cite{majumder2019dialoguernn} employs Recurrent Neural Networks (RNNs) to capture the speaker identity as well as the historical context and the emotions of past utterances to capture the nuances of conversation dynamics. ConGCN \cite{zhang2019modeling} utilizes a Graph Convolutional Network (GCN) to represent relationships within a graph that incorporates both context and speaker information of multiple conversations. MMGCN \cite{hu2021mmgcn} also proposes a GCN-based approach, but captures representations of a conversation through a graph that contains long-distance flow of information as well as speaker information. DialogueTRM \cite{mao2021dialoguetrm} focuses on modeling both local and global context of conversations to capture the temporal and spatial dependencies. DAG-ERC \cite{shen2021directed} studies how conversation background affects information of the surrounding context of a conversation. MMDFN \cite{hu2022mm} proposes a framework that aims to enhance integration of multimodal features through dynamic fusion. EmoCaps \cite{li2022emocaps} introduces an emotion capsule that fuses information from multiple modalities with emotional tendencies to provide a more nuanced understanding of emotions within a conversation. UniMSE \cite{hu2022unimse} seeks to unify ERC with multimodal sentiment analysis through a T5-based framework. GA2MIF \cite{li2023ga2mif} introduces a two-stage multimodal fusion of information from a graph and an attention network. FacialMMT \cite{zheng2023facial} focuses on extracting the real speaker's face sequence from multi-party conversation videos and then leverages auxiliary frame-level facial expression recognition tasks to generate emotional visual representations.

\subsection{Error Analysis}
\label{sec:appendix_6}
\begin{figure}[hbt!] 
\centering

\begin{center}

\includegraphics[width=1.0\linewidth]{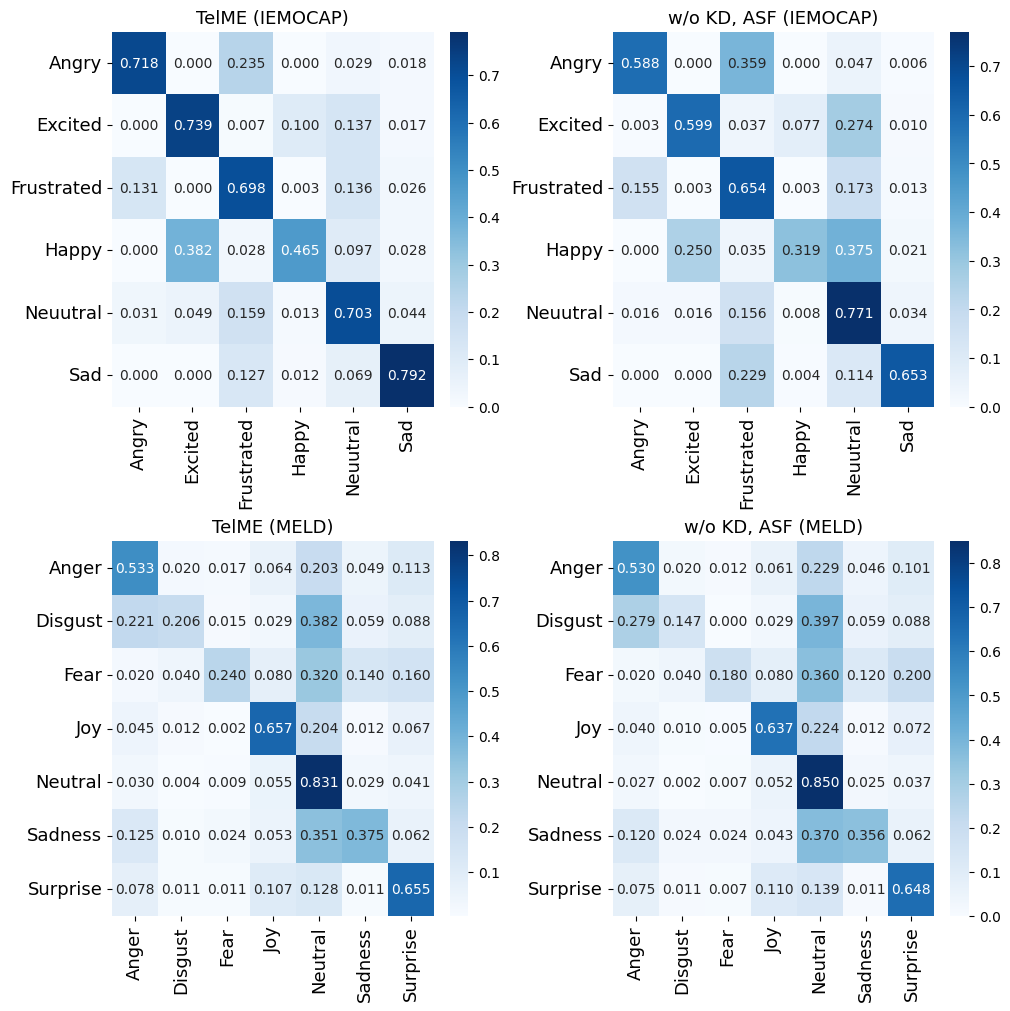}

\end{center}

\caption{Confusion Matrices on IEMOCAP and MELD}

\label{fig:appendix_A6}
\end{figure}
Figure \ref{fig:appendix_A6} shows the normalized confusion matrices of the TelME and the understated model for two datasets. We can evaluate the quality of the emotion prediction through the confusion matrix. TelME shows better True Positive results in almost all emotion classes. This suggests that TelME is extracting and fusing finer-grained features to infer emotions without bias. TelME better classifies similar emotions compared to the understated model(e.g., excited and happy, angry and frustrated). However, the result of misclassifying happy as exciting is a little high. This result is due to the lowest percentage of happy in IEMOCAP with unbalanced classes. Even in the case of MELD, the emotion in which most emotion classes are misclassified is neutral, with the highest count. We can observe a similar misclassification tendency in other research ~\citep{chudasama2022m2fnet, hu2023supervised} as well. Hence, we suspect that the cause of misclassification is not a problem with the method we proposed but rather stems from a class imbalance issue.

\subsection{Results of Random Seed Numbers}
\label{sec:appendix_7}
\begin{table}[hbt!]
\resizebox{\columnwidth}{!}{%
\begin{tabular}{cccc}
\hline
   SEED      & MELD        & IEMOCAP  \\ \hline
   0         & 67.27       &  70.50   \\
   1          & 67.41       &  70.69   \\
 1234        & 67.44       &  70.21    \\
 2023         & 67.24       &  69.95    \\
  42           & 67.37       &  70.48    \\ \hline \hline
mean       & 67.35 & 70.37          \\ 
standard deviation & 0.0781 & 0.2581      \\ \hline
\end{tabular}%
}
\caption{Performance of the full framework for five random seeds}
\label{tab:Table10}
\end{table}

We report all outcomes based on the seed number 42 following previous studies~\citep{lee2021compm,song2022supervised, hu2022unimse}. However, to validate TelME's robustness to randomness, we present experiment results with different seed numbers in Table~\ref{tab:Table10}. The results in Table 10 demonstrate that the performance of TelME is robust to seed variations.

\subsection{Utility of TelME}

\begin{table}[hbt!]
\centering
\resizebox{\columnwidth}{!}{%
\begin{tabular}{ccccc}
\hline
TelME & Label & Text & Audio & Visual \\ \hline \hline
\textbf{anger}     & \textbf{anger}     & disgust    & \textbf{anger}     & neutral      \\
\textbf{neutral}     & \textbf{neutral}     & surprise    & \textbf{neutral}    & \textbf{neutral}      \\
\textbf{joy}     & \textbf{joy}     & surprise    & anger     & neutral      \\
\textbf{anger}     & \textbf{anger}     & surprise    & neutral     & neutral      \\
\textbf{joy}     & \textbf{joy}     & disgust    & \textbf{joy}     & \textbf{joy}      \\
\textbf{sadness}     & \textbf{sadness}     & fear    & neutral      & neutral       \\
\textbf{joy}     & \textbf{joy}     & surprise    & \textbf{joy}     & anger      \\
\textbf{neutral}     & \textbf{neutral}     & sadness    & \textbf{neutral}     & sadness      \\ \hline \hline
\end{tabular}%
}
\caption{inference results of each unimodal model and TelME on MELD}
\label{tab:Table11}
\end{table}

Table~\ref{tab:Table11} presents a segment of the ground truth label from the MELD dataset, along with inference outcomes of each unimodal model (Text teacher, non-verbal students) and TelME. The results indicate that student models can make different judgments than the text teacher even after knowledge distillation. Moreover, the final decision of TelME, supported by complementary information from non-verbal modalities, might diverge from the prediction of the text teacher, rectifying any inaccuracies. This implies that TelME utilizes multimodal information instead of heavily depending on any of the three modalities.

\begin{table*}[hbt!]
\centering
\resizebox{\textwidth}{!}{%
\begin{tabular}{cccccccc}
\hline
\multirow{2}{*}{Models} & \multicolumn{7}{c}{IEMOCAP: Emotion Categories}                \\
                        & Happy & Sad   & Neutral & Anger & Excited & Frustrated & F1    \\ \hline
DialogueRNN~\citep{majumder2019dialoguernn}             & 33.18 & 78.80 & 59.21   & 65.28 & 71.86   & 58.91      & 62.75 \\
MMGCN~\citep{hu2021mmgcn}                   & 42.34 & 78.67 & 61.73   & 69.00 & 74.33   & 62.32      & 66.22 \\
DialogueTRM~\citep{mao2021dialoguetrm}             & 48.70 & 77.52 & 74.12   & 66.27 & 70.24   & 67.23      & 69.23 \\
DAG-ERC~\citep{shen2021directed}                 & -     & -     & -       & -     & -       & -          & 68.03 \\
MM-DFN~\citep{hu2022mm}                  & 42.22 & 78.98 & 66.42   & 69.77 & 75.56   & 66.33      & 68.18 \\
M2FNet~\citep{chudasama2022m2fnet}                  & -     & -     & -       & -     & -       & -          & 69.86 \\
EmoCaps~\citep{li2022emocaps}                 & \textbf{71.91} & \textbf{85.06} & 64.48   & 68.99 & \textbf{78.41}   & 66.76      & \textbf{71.77} \\
UniMSE~\citep{hu2022unimse}                  & -     & -     & -       & -     & -       & -          & 70.66 \\
GA2MIF~\citep{li2023ga2mif}                  & 46.15 & 84.50 & \textbf{68.38}   & \textbf{70.29} & 75.99   & 66.49      & 70.00 \\ \hline \hline
TelME                   & 49.46 & 83.48 & 67.42   & 68.49 & 77.38   & \textbf{68.63}      & 70.48 \\ \hline
\end{tabular}%
}
\caption{Performance comparisons on IEMOCAP (6-way)}
\label{tab:Table12}
\end{table*}

\subsection{Performance comparisons on IEMOCAP}
Table~\ref{tab:Table12} shows a performance comparison of our approach and other approaches on IEMOCAP dataset. 
\end{document}